\documentclass{article}
\usepackage{ijcai19}

\usepackage{booktabs}
\usepackage{courier}
\usepackage{epsfig}
\usepackage{times}
\usepackage{soul}
\usepackage{url}
\usepackage[hidelinks]{hyperref}
\usepackage[utf8]{inputenc}
\usepackage[small]{caption}
\usepackage{graphicx}
\usepackage{helvet}
\usepackage{amsmath}
\usepackage{amssymb}
\usepackage{threeparttable}
\usepackage{multirow}
\usepackage{subfigure} 
\usepackage[linesnumbered,ruled,vlined]{algorithm2e}
\usepackage{algpseudocode}  
\usepackage{color}
\usepackage{array}

\title{Graph Attribute Aggregation Network with Progressive Margin Folding}

\author{
Penghui Sun$^1$
\and
Jingwei Qu$^1$\and
Xiaoqing Lyu$^{1}$\and
Hainbin Ling$^2$\And
Zhi Tang$^1$
\affiliations
$^1$Peking University
$^2$Temple University\\
\emails
\{sph, qujingwei, lvxiaoqing\}@pku.edu.cn,
hbling@temple.edu,
tangzhi@pku.edu.cn
}

\begin{document}

\maketitle

\begin{abstract}
Graph convolutional neural networks (GCNNs) have been attracting increasing research attention due to its great potential in inference over graph structures. However, insufficient effort has been devoted to the aggregation methods between different convolution graph layers. In this paper, we introduce a graph attribute aggregation network (GAAN) architecture. Different from the conventional pooling operations, a graph-transformation-based aggregation strategy, progressive margin folding, PMF, is proposed for integrating graph features. By distinguishing internal and margin elements, we provide an approach for implementing the folding iteratively. And a mechanism is also devised for preserving the local structures during progressively folding. In addition, a hypergraph-based representation is introduced for transferring the aggregated information between different layers. Our experiments applied to the public molecule datasets demonstrate that the proposed GAAN outperforms the existing GCNN models with significant effectiveness.
\end{abstract}

\begin{figure*}[!t]
	\centering
	\includegraphics[width=2\columnwidth]{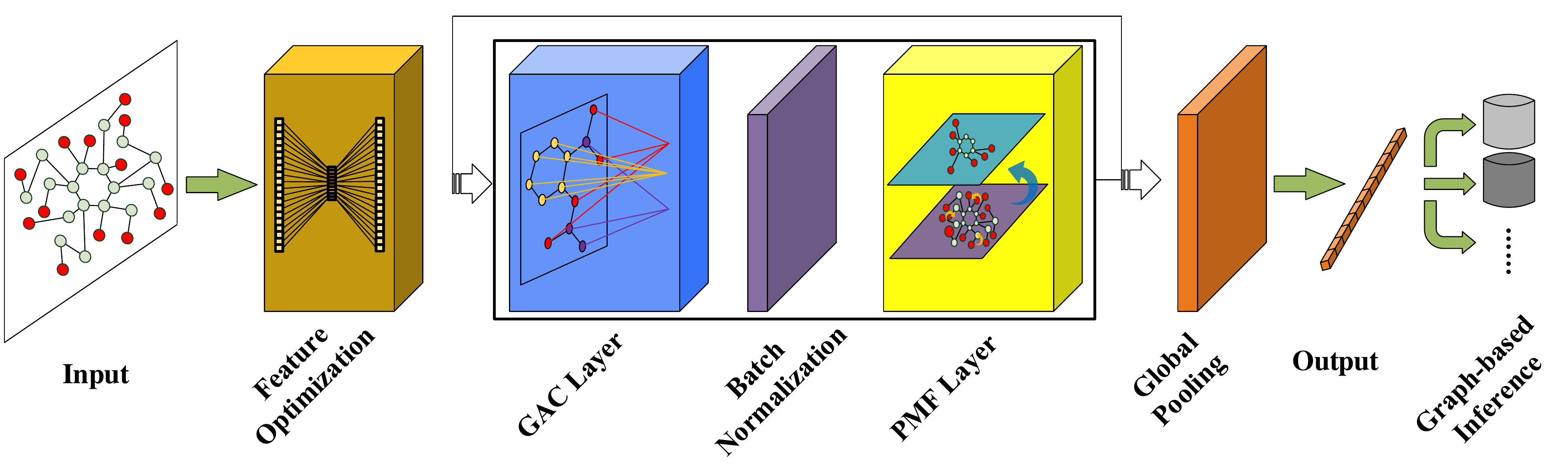}
	\caption{Network architecture of GAAN.}
	\label{fig-architecture}
\end{figure*}

\section{Introduction}
Convolutional neural networks (CNNs) have brought significant improvements in many areas such as speech recognition, image classification, video understanding, etc. CNN is mainly applied to regular grid-structured data, where convolution operation can be effectively implemented. However, in other applications, e.g., point clouds, molecules, and social networks, data represented as irregular structures or in non-Euclidean domains, the conventional grid-based convolution is hardly applicable to graphs directly. Such data are usually structured as graphs to depict characteristics of entities and relations among them. Due to the requirements of the locality, stationarity, compositionality of data representation, generalization of CNN from grid-structured data to irregular graphs, i.e., graph convolutional neural networks (GCNNs) \cite{bruna2013spectral,henaff2015deep}, is highly desired.

In our view, there are two key components in a GCNN model, the convolution kernels capturing information on both vertices and edges in a graph at the current layer, and the aggregation method transferring the information between different layers. Most researches about GCNN focus on the innovations of the former, and generally adopt the conventional pooling strategies, e.g. max pooling and graph coarsening, to implement the information aggregation. However, the pooling (or equivalent) strategies in the existing GCNN models are usually locality insensitive, which are inappropriate for graph structures where all vertices are not created equally. For example, the molecular scaffold is of vital importance in a molecular graph. Consequently, locality insensitive pooling strategies may destroy important graph structure too early when abstracting a graph progressively.

Improving the performance of GCNNs faces many challenges. For instances, to explore an effective aggregation approach for transferring information between the convolution layers, which is different from the ordinary pooling methods, is more difficult than designing the convolution kernels. Most abstracting methods proposed for image processing can hardly achieve a satisfying result for graphs, which lead us to pay more attention to those graph-transform-based approaches. 
Although some graph algorithms already have been introduced into the deep learning neural networks, e.g. graph coarsening, most of them are not sensitive to the local structures of a graph and could not detect or maintain the corresponding meaningful information. An important problem is how to avoid destroying useful graph structures too early when abstracting a graph.

To address the issues discussed above, in this paper we propose a graph attribute aggregation network (GAAN) architecture for graph-based inference applications. To effectively capture rich attribute information on both vertices and edges, we design a graph attribute convolution (GAC) operation. To maintain important internal structure during progressively abstracting, we propose a progressive margin folding (PMF) operation. 
Unlike conventionally used pooling operations in CNNs or GCNNs, PMF is locality sensitive and aims to preserve important structures in the interior of a graph. 
To summarize, we make the following key contributions:

\begin{itemize}
\item Different from the conventional pooling operations, a graph-transformation-based aggregation strategy, PMF, is proposed for integrating graph features in the convolution layers at different scales.
\item By dynamically distinguishing the elements into internal and margin ones, we provide a possible approach to implement the folding iteratively.
\item A mechanism is devised for preserving the possibly meaningful local structures at the appropriate scales during progressively folding.
\item A hypergraph-based representation method is introduced for transferring the aggregated information between different convolution layers.
\end{itemize}

With the proposed GAAN model, we implement several graph-based inference applications. In this paper, we provide the experiments about molecular property prediction including classification and regression tasks. The evaluation, conducted on publicly available benchmarks, demonstrates the advantages of the proposed model over the existing GCNNs.

The rest of this paper is organized as follows. Section 2 summaries the related work in the field of GCNNs. Section 3 presents the details of the proposed GAAN architecture, including GAC and PMF. Section 4 reports our experimental results in details. Section 5 concludes this paper.

\section{Related Work}
\label{sec2}
Existing GCNN algorithms can be roughly divided into two categories: spectral domain graph convolution and spatial domain graph convolution.

A mathematically sound definition of convolution operator makes use of the spectral graph theory \cite{chung1997spectral}.
As one of the first methods to generalize the CNNs to graph-structured data, \cite{bruna2013spectral} designed the convolution operator by a point-wise product in the spectrum domain according to the convolution theorem. Later, the spectrum filtering based methods using Chebyshev polynomials and Cayley polynomials are proposed in \cite{defferrard2016convolutional} and \cite{levie2017cayleynets} respectively. \cite{kipf2016semi} simplified the spectrum method \cite{defferrard2016convolutional} with a first-order approximation to the Chebyshev polynomials as the graph filter spectrum, which requires much less training parameters. \cite{henaff2015deep} developed an extension of the spectral networks to incorporate a graph estimation procedure. By transferring the intrinsic geometric information learned in the source domain, \cite{lee2017transfer} constructed a model for a new but related task in the target domain without collecting new data nor training a new model from scratch. 
\cite{feng2018hypergraph} designed a hyper convolution operation is designed to handle the data correlation during representation learning. \cite{gilmer2017neural} reformulated some existing models into a single common framework called Message Passing Neural Networks (MPNNs) and explored additional novel variations within this framework. Besides, \cite{Li2018Adaptive} constructed unique residual Laplacian matrix for graph-structured data and learned distance metric for graph update. 

On the other hand, some researchers worked on designing feature propagation models in the spatial domain for GCNNs. \cite{duvenaud2015convolutional} proposed a CNN model that operates directly on graphs for learning molecular fingerprints. \cite{niepert2016learning} also proposed a framework for learning CNN for arbitrary graphs. Their framework firstly selects a vertex sequence from a graph via a graph labeling procedure. Then, it assembles and normalizes a local neighborhood graph used as receptive fields. The graph max-pooling and graph-gathering layers are designed in \cite{altae2017low} for increasing the size of downstream convolutional layer receptive fields without increasing the number of parameters. \cite{simonovsky2017dynamic} formulated a convolution-like operation on graph signals performed in the spatial domain where filter weights are conditioned on edge labels (discrete or continuous) and dynamically generated for each specific input sample, and they applied the edge-conditioned convolution (ECC) to point cloud classification. 
PointNet \cite{qi2017pointnet} and PointNet++ \cite{qi2017pointnet++}, which specially work on point cloud, respect well the permutation invariance of points in the input. \cite{wang2018local} used spectral graph convolution on a local graph, combined with a recursive clustering and pooling strategy. 

In the existing GCNN models, the pooling (or equivalent) strategies are mainly inspired by conventional CNNs or graph clustering algorithms. Some GCNNs \cite{kearnes2016molecular,altae2017low} simply return the maximum activation across a receptive filed, which is analogous to the max-pooling operation in CNNs. In addition, by clustering vertices of a graph in multiple layers, \cite{defferrard2016convolutional,simonovsky2017dynamic} utilize a graph coarsening strategy to achieve a multi-scale clustering of the graph.


\section{Graph Attribute Aggregation Network}
\label{sec3}
We first present the overall GAAN architecture, and then elaborate on GAC and PMF operations.

\subsection{GAAN Architecture}
Our GAAN architecture, as shown in Figure \ref{fig-architecture}, consists of the following types of neural network layers: feature optimization layer, GAC layer, batch normalization layer, PMF layer, and global pooling layer.

Firstly, a feature optimization layer implemented with an auto-encoder network is adopted to encode raw sparse features of the input graph with arbitrary structures and sizes. 
Then the features of the graph in different depths are extracted with multiple combos of a GAC layer, a batch normalization layer \cite{Ioffe2015Batch}, and a PMF layer. In a GAC layer, the graph convolutional kernels are implemented with specific finite attributes for practical applications. 
Next the information aggregation of the graph is accomplished hierarchically by the PMF layer, and the newly generated graph is transferred to the next layer. Besides, the similar layer combos are added with different parameter configuration in order. Finally, a global pooling layer averages all the feature vectors of the whole graph. The output of this pooling layer is adopted as the graph-level representation. 

\subsection{Graph Attribute Convolution}
For a graph $G=(V, E)$, where $V$ is a finite set of $n=|V|$ vertices, $E \subseteq V\times V$ is a set of $m=|E|$ edges, the graph features are defined as the following mappings,
\begin{eqnarray}
    X^l_V: V \mapsto \mathbb{R}^{d_V^l} \ ,
    \quad X^l_E: E \mapsto \mathbb{R}^{d_E^l}, \nonumber
\end{eqnarray}
 where $X^l_V$ and $X^l_E$ correspond to features on vertices and edges, respectively; $l \in \{0, ..., l_{max}\}$ indicate the layer in a feed-forward neural network; and $d_V^l$ and $d_E^l$ are corresponding feature dimensions. 
 The $0$-th layer ($l=0$) is the original input with the associated features $X^0_V$ and $X^0_E$.
 
\begin{algorithm} [!b]
	\caption{Graph Attribute Convolution (GAC)}
	\label{alg-conv}
	\KwIn{Graph Batch $B^{l-1} = \{G_1^{l-1}, ..., G_s^{l-1}\}$ \\
	\qquad \quad \textbf{Weight} ${\rm\bold{W}}_V^l$ 
	${\rm\bold{W}}_E^l$ 
	\qquad \textbf{Bias} ${\rm\bold{b}}_V^l$ 
	 ${\rm\bold{b}}_E^l$ 
	}
	\KwOut{Feature Graphs $B^l = \{G_{1}^l, ..., G_{s}^{l}\}$}
	
	\For{$G_k^{l-1} \in B^{l-1}$}
	{
		${\rm\bold{A}}_V, {\rm\bold{A}}_E \leftarrow$ GetAttributeValue($V_k^{l-1}$, $E_k^{l-1}$) \\
		${\rm\bold{V}}, {\rm\bold{E}} \leftarrow$ Classify($V_k^{l-1}$, ${\rm\bold{A}}_V$, $E_k^{l-1}$, ${\rm\bold{A}}_E$) \\
		\For{$V_i \in {\rm\bold{V}}$}
		{
			$X^l_V(V_i) \leftarrow X^{l-1}_V(V_i)W_V^l(i) + b_V^l(i)$ \\
		}
		\For{$E_j \in {\rm\bold{E}}$}
		{
			$X^l_E(E_j) \leftarrow X^{l-1}_E(E_j)W_E^l(j) + b_E^l(j)$ \\
		}
		$X^l \leftarrow \lambda X^l_V + (1 - \lambda)X^l_E$ \\
		$G_k^l \leftarrow \sigma(G_k^{l-1}(X^l))$ // activation function
	}
	return $B^l$
\end{algorithm}

To deal with topologically weak structures and reduce the burden of reliance on fixed kernel design, e.g. vertex degrees, we devise a \emph{graph attribute convolution} (GAC) operation by integrating attribute information, as summarized in Algorithm~\ref{alg-conv}. 
Specifically, we select an intrinsic vertex attribute set ${\rm\bold{A}}_V$ with discrete and finite values ${\rm\bold{A}}_V = \{A_V^1,..., A_V^p\}$, and similarly, an intrinsic edge attribute set ${\rm\bold{A}}_E$ with ${\rm\bold{A}}_E = \{A_E^1, ..., A_E^q\}$. Consequently, all vertices in $G$ can be classified into finite groups ${\rm\bold{V}} = \{V_1, ..., V_p\}$, such that $V_i \subseteq V$ with attribute value $A_V^i$. Similarly, edges are categorized by ${\rm\bold{A}}_E$, i.e, ${\rm\bold{E}} = \{E_1, ..., E_q\}$.
Then the convolutional feature $X^l_V(v)$ on a vertex $v \in V_i$ is learned by updating the weight $W_V^l(i) \in \mathbb{R}^{d_V^{l-1} \times d_V^l}$ and bias $b_V^l(i) \in \mathbb{R}^{d_V^l}$ corresponding to its category:
\begin{equation}
\label{eq1}
X^l_V(V_i) = X^{l-1}_V(V_i)W_V^l(i) + b_V^l(i)
\end{equation}
Similar convolution operation is conducted on an edge $e \in E_j$:
\begin{equation}
\label{eq2}
X^l_E(E_j) = X^{l-1}_E(E_j)W_E^l(j) + b_E^l(j)
\end{equation}
Then GAC is conducted by fusing the learned features:
\begin{equation}
\label{eq3}
    X^l \leftarrow \lambda X^l_V + (1 - \lambda)X^l_E
\end{equation}


where $\lambda$ is a weight coefficient. We adopt the same number of convolutional kernels to learn the features of vertices and edges, thus they have the same number of output channels of corresponding feature map. Afterward, a non-linear activation function is applied to the convolutional results. 

\subsection{Progressive Margin Folding}
As discussed before, most convolutional layers in the existing GCNNs generally adopt the off-the-shelf pooling strategies lacking the sensitivity of local structures, or specifically, they mainly focus on evolving the features on vertices without changing the underlying graph. To address this issue, we devise a graph-transformation-based strategy, PMF, which aims to aggregate iteratively the information between different layers based on the technique of graph abstraction.

First of all, we distinguish the vertices into internal and margin nodes and propose an algorithm for searching marginal structures. Also preparing for the further folding, we detect the necessary and probably meaningful local structures, such as cycles for molecule graphs (detected with RDKit).  Secondly, in each iteration of PMF, we fold inward almost all marginal vertices along the graph periphery and re-examine the marginal structures for the next iteration. It should be noted that we take into consideration not only the marginal vertices but also the local structures including judging the time to process them. Moreover, to finish transferring the complete information from a lower layer to an upper layer, we design a hypergraph-based method representing the current folding results.

For an input graph $G=(V,E)$, PMF constructs a pyramid of $h_{max}$ progressively-abstracted graphs, denoted by $G^{h, l_h} = (V^{h, l_h}, E^{h, l_h})$, $h \in \{0, ..., h_{max}\}$, where $l_h$ is the layer index, and $G^{0, 0}=G$ is the input graph. The main procedure of PMF is shown in Algorithm 2. 
\begin{algorithm}[!b]
	\caption{Progressive Margin Folding (PMF)}
	\label{alg-contraction}
	\KwIn{Feature graph batch $B^{l_{h-1}} = \{G_1^{h-1, l_{h-1}}, ..., G_s^{h-1, l_{h-1}}\}$
	}
	\KwOut{Abstracted graphs $B^{l_h} = \{G_1^{h, l_h}, ..., G_s^{h, l_h}\}$}
	\For{$G_k^{h-1, l_{h-1}} \in B^{l_{h-1}}$}
	{
		$V_M^{h-1, l_{h-1}} \leftarrow$ GetMarginalStructure($G_k^{h-1, l_{h-1}}$) \\
		\For{$v \in V_M^{h-1, l_{h-1}}$}
		{
			$u \leftarrow$ GetNeighbor($v$) \\
			$X^{h, l_h}_V(u) \leftarrow X^{h-1, l_{h-1}}_V(u) + \alpha X^{h-1, l_{h-1}}_V(v) + \beta X^{h-1, l_{h-1}}_E\big((u, v)\big)$ \\
			$u \leftarrow u \cup v$ \\
      		}
      		\If{branch of ring$ \leq $1}
      		{
      			ring collapsing \\
      		}
      		hypergraph generating
	}
	return $B^{l_h}$
\end{algorithm}

\subsubsection{Marginal Structure Search}
\begin{figure}[!t]
	\centering
	\subfigure[]{\label{fig3a}\includegraphics[width=0.3\columnwidth]{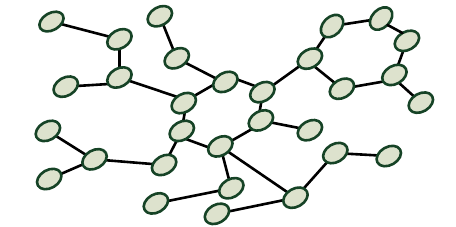}}
	\hspace{1ex}
	\subfigure[]{\label{fig3b}\includegraphics[width=0.3\columnwidth]{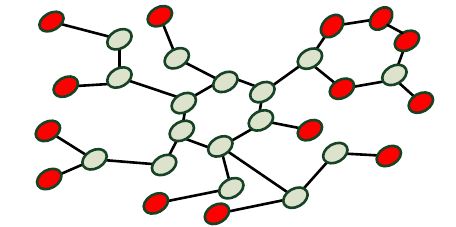}}
	\hspace{1ex}
	\subfigure[]{\label{fig3c}\includegraphics[width=0.3\columnwidth]{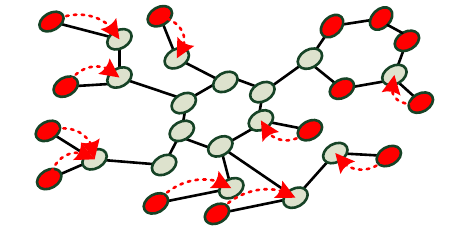}}
	\subfigure[]{\label{fig3d}\includegraphics[width=0.27\columnwidth]{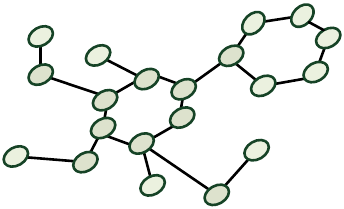}}
	\hspace{4ex}
	\subfigure[]{\label{fig3e}\includegraphics[width=0.27\columnwidth]{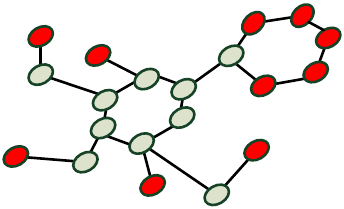}}
	\hspace{4ex}
	\subfigure[]{\label{fig3f}\includegraphics[width=0.27\columnwidth]{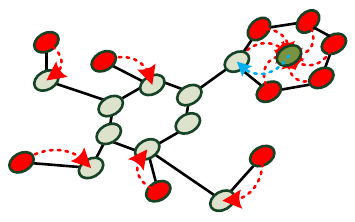}}
	\subfigure[]{\label{fig3g}\includegraphics[width=0.25\columnwidth]{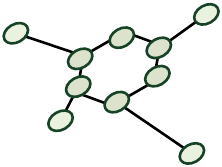}}
	\hspace{4ex}
	\subfigure[]{\label{fig3h}\includegraphics[width=0.25\columnwidth]{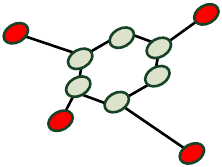}}
	\hspace{4ex}
	\subfigure[]{\label{fig3i}\includegraphics[width=0.25\columnwidth]{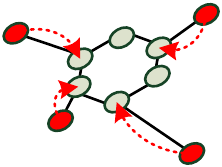}}
	\caption{Sample steps of PMF. (a) Original graph $G^{0, 0}$; (b) Marginal structure $V_M^{0, 0}$; (c) Folding $V_M^{0, 0}$. (d) $G^{1, l_1}$; (e) Marginal structure $V_M^{1, l_1}$; (f) Folding $V_M^{1, l_1}$; (g) $G^{2, l_2}$; (h) Marginal structure $V_M^{2, l_2}$; (i) Folding $V_M^{2, l_2}$.}
	\label{fig-pmf}
\end{figure}

To implement the folding operation, the marginal structures need to be searched first. The marginal elements include marginal vertices (shown as the red nodes in Figure \ref{fig-pmf} (b)) and marginal edges. In consideration of the meaningful local structures of a graph, such as the cyclic and acyclic structures, some marginal elements may be embedded in a local structure not just dangling as a leaf. Thus different methods were designed for detecting different marginal elements.  In brief, for the acyclic part of a graph, a vertex is identified as marginal vertex if the degree of the vertex is 1 and an edge is identified as a marginal edge if the edge connects the marginal vertex. For the part of a ring structure, a vertex is identified as marginal vertex if its degree is 2 and an edge is identified as a marginal edge if the edge connects two marginal vertices of the ring.

It is worth noting that the concept of marginal vertices is a relative concept. The marginal vertices and edges are selected and updated dynamically in each folding iteration. An internal vertex when all its neighbor are folded in the current iteration may become a marginal vertex in the next iteration of the folding procedure. By folding the marginal vertices iteratively, it can be guaranteed to decrease the graph size.

\subsubsection{Dynamic Folding Inward}
The folding procedure starts with the input graph. After accomplishing the convolutional computation, we sum and transfer the information of marginal vertices into their inner neighbors. 
For each marginal vertex $v\in V^{h-1, l_{h-1}}$ with associated edge $(u, v) \in E^{h-1, l_{h-1}}$, we transfer the features of $v$ and $(u, v)$ into its neighbor $u$:
\begin{equation}
\label{eq5}
\begin{aligned}
X^{h, l_h}_V(u) \leftarrow& X^{h-1, l_{h-1}}_V(u) + \alpha X^{h-1, l_{h-1}}_V(v) + \\& \beta X^{h-1, l_{h-1}}_E\big((u, v)\big)
\end{aligned}
\end{equation}
where the learnable weight parameters $\alpha$ and $\beta$ are mainly used to balance the impact of the current vertex and edge on the feature aggregation. Then the layer with index $l_h$ aggregates vertex features $X^{h-1, l_{h-1}}_V$ and edge features $X^{h-1, l_{h-1}}_E$ into $X^{h, l_h}_V$. For example, in the first step, the embedding of marginal vertex is folded into the marginal edge. In the next step, the embedding of marginal edge by graph attribute convolution is folded into the inner marginal vertex,  indicated as red nodes and arrows in Figure \ref{fig-pmf} (c), (f), and (i). 
Finally, a new graph $G^{h, l_h}$ is constructed by trimming the marginal vertices and their incident edges in $G^{h-1, l_{h-1}}$. Graph $G^{h, l_h}$ is progressively delivered to the next layer for further convolution and folding. 

During the folding, the local structures are processed simultaneously. 
Taking the ring shown in the top-right of Figure 3 for an example, it depends on a judgment of timing for processing. Although some vertices in this ring are marginal from the beginning, it should wait for the situation that all of them have been folded except one node connecting with the main part of a graph to collapse the ring into a surrogate vertex (Figure \ref{fig-pmf} (f)). In other words, a ring is collapsed if there is only one branch or no branch that connects to its nodes. As we adopt the detection algorithm for cycles in advance, more refined decisions are integrated for other connection modes of rings. By such a reduction for local structures in a graph, we could not only obtain the abstraction representation at the proper layer but also simplify the large graph efficiently. 

It is obviously that the folding algorithm is different from pruning algorithm. The important information of marginal vertices is transferred to inner vertices instead of discarding. As a consequence, the PMF operation can be applied to general-purpose graphics and can be modified easily for other specific graphs from many other modalities or domains.

\subsubsection{Information Transfer with Hierarchical Representation}
Each folding layer aggregates graph features and transfers them from a lower layer to an upper layer.  Both vertex folding and ring collapsing need a new representation method for transferring the results of aggregation of the current layer to the upper layer.  To maintain the integrity of the information and also to reduce the calculation cost, we introduce a hypergraph-based method for the description. When an iteration of folding in PMF is performed, a hypergraph is dynamically constructed for the updated internal vertices, which contains not only the information from the newly folded marginal vertices and edges but also the information of possibly generated surrogate element representing a  local structure.  In short, we aggregate features of original vertices and edges into the corresponding hypervertices:

\begin{equation}
\label{eq6}
X^l_V(R) \leftarrow \sum_{v \in R_V}\omega_v X^l_V(v) + \sum_{e \in R_E}\theta_e X^l_E(e)
\end{equation}
where $R$ denotes the collapsed ring, the weight parameters $\omega$ and $\theta$ are learned during the model training. Then a hypergraph is constructed by inserting hyperedges according to topological relations between these hypervertices.

In the final layer, the global representation of the whole graph is obtained by integrating the features in the different graphs of the pyramid. The folding in each PMF layer is a global operation for the whole graph, thus PMF avoids the risk of over-localization. Besides, PMF possesses high generality and can be applied to various topology structures of graphs with preserving the completeness of key substructures. 

The complexity of PMF is dependent on the number of margin vertices, which is $O(n)$ in the worst case where all vertices are marginal except for the center vertex, i.e., a star-like structure of the underlying graph. 


\section{Experiments}
\label{sec4}
The GAAN model is evaluated 
in the classification and regression prediction of molecular properties.

\subsection{Multi-task Classification and Regression}
Prediction of molecular properties plays an important role in virtual drug screening, material design, etc. Obtaining accurate properties typically requires high quality feature representation for molecules. To demonstrate the superiority of our GAAN model on molecular property prediction, we compare it with four state-of-the-art GCNNs: ($\romannumeral1$) the first spectral CNN (SCNN) \cite{bruna2013spectral} with linear B-spline interpolated kernel; ($\romannumeral2$) neural fingerprint (NFP) \cite{duvenaud2015convolutional} which is the cutting-edge neural network for molecules; ($\romannumeral3$) the extension to SCNN, the spectral CNN implemented with Chebyshev polynomials-based spectral filter (ChebNet) \cite{defferrard2016convolutional}; and ($\romannumeral4$) the adaptive graph convolutional neural network (AGCN) \cite{Li2018Adaptive} for learning adaptive graph topology structure.

\subsubsection{Datasets}
\begin{table}[!t]
	\centering
	\caption{Datasets details.}
	\label{tab-datasets}
	\begin{tabular}{m{1.3cm}<{\centering} m{1.7cm}<{\centering} m{1.7cm}<{\centering} m{0.7cm}<{\centering} m{0.8cm}<{\centering}}
		\toprule
		Field & Dataset & Task type & Tasks & Graphs \\
		\midrule
		\multirow{4}{*}{Physiology} & ClinTox & \multirow{4}{*}{Classification} & 2 & 1491 \\
		& SIDER & & 27 & 1427      \\
		& Tox21 & & 12 & 8014      \\
		& ToxCast & & 617 & 8615         \\ \hline
		\multirow{4}{*}{\begin{tabular}[c]{@{}l@{}}Physical\\ Chemistry\end{tabular}} & ESOL & \multirow{4}{*}{Regression} & 1 & 1128 \\
		& Lipophilicity & & 1 & 4200      \\
		& NCI & & 60 & 19126      \\
		& FreeSolv & & 1 & 643      \\
		\bottomrule
	\end{tabular}
\end{table}

\begin{table*}[!t]
	\centering
	\caption{Mean ROC-AUC scores on ClinTox, SIDER, Tox21, and ToxCast datasets.}
	\label{tab-classification}
	\renewcommand{\multirowsetup}{\centering}
	\begin{tabular}{ccccccccc}
		\toprule
		\multirow{2}{*}{Model}  & \multicolumn{2}{c}{ClinTox} & \multicolumn{2}{c}{SIDER} & \multicolumn{2}{c}{Tox21} & \multicolumn{2}{c}{ToxCast} \\
		\cmidrule{2-9}
		& Validation   & Testing & Validation    & Testing  & Validation   & Testing & Validation  & Testing  \\
		\midrule
		SCNN     & 0.7896   & 0.7069  & 0.5806   & 0.5642  & 0.7105       & 0.7023    & 0.6479   & 0.6496   \\
		NFP      & 0.7356    & 0.7469  & 0.6049   & 0.5525  & 0.7502   & 0.7341  & 0.6561   & 0.6384   \\
		ChebNet     & 0.8303   & 0.7573  & 0.6085   & 0.5914   & 0.7540   & 0.7481  & 0.6914   & 0.6739   \\
		AGCN   & 0.9267   & 0.8678  & 0.6112   & 0.5921    & 0.7947   & 0.8016   & 0.7227  & 0.7033   \\
		\midrule
		GAAN & \textbf{0.9384} & \textbf{0.8877}& \textbf{0.6650} & \textbf{0.6575} & \textbf{0.8209}  & \textbf{0.8387} & \textbf{0.7376} & \textbf{0.7293}  \\
		\bottomrule
	\end{tabular}
\end{table*}

\begin{table*}[!t]
	\centering
	\caption{Mean and standard deviation of RMSE on ESOL, Lipophilicity, NCI, and FreeSolv datasets. }
	\label{tab-regression}
	\begin{tabular}{ccccc}
		\toprule
		Model & ESOL & Lipophilicity  & NCI  & FreeSolv \\
		\midrule
		SCNN  & 0.4222 $\pm$ 8.38e-2    & 0.7516 $\pm$ 8.42e-3 & 0.8695 $\pm$ 3.55e-3 & 2.0329 $\pm$ 2.70e-2 \\
		NFP  & 0.4955 $\pm$ 2.30e-3    & 0.9597 $\pm$ 5.70e-3 & 0.8748 $\pm$ 7.50e-3 & 3.4082 $\pm$ 3.95e-2 \\
		ChebNet  & 0.4665 $\pm$ 2.07e-3    & 1.0459 $\pm$ 3.92e-3 & 0.8717 $\pm$ 4.14e-3 & 2.2868 $\pm$ 1.37e-2 \\
		AGCN & 0.3061 $\pm$ 5.34e-3    & 0.7362 $\pm$ 3.54e-3 & 0.8647 $\pm$ 4.67e-3 & 1.3317 $\pm$ 2.73e-2 \\
		\midrule
		GAAN  & \textbf{0.2936 $\pm$ 4.87e-3} & \textbf{0.6045 $\pm$ 4.10e-3} & \textbf{0.8472 $\pm$ 2.20e-3} & \textbf{1.0568 $\pm$ 2.86e-2} \\
		\bottomrule
	\end{tabular}
\end{table*}

Multiple public molecular graph datasets \cite{wu2018moleculenet} (as summarized in Table \ref{tab-datasets}) in different fields are adopted to evaluate our GAAN model for thoroughly studying the performance. The ClinTox dataset addresses clinical drug toxicity and consists of two classification tasks: clinical trial toxicity and FDA approval status. The side effect resource (SIDER) \cite{kuhn2015sider} is a dataset of marketed drugs and adverse drug reactions.
The toxicology in the 21st century (Tox21) for measuring the toxicity of compounds was used in the 2014 Tox21 Data Challenge. It contains qualitative toxicity measurements for 8,014 compounds.
The ToxCast \cite{richard2016toxcast} dataset provides qualitative toxicology data of more than 600 experiments on 8615 compounds.

ESOL \cite{delaney2004esol} is a dataset consisting of water solubility data for 1,128 compounds. Lipophilicity provides experimental results of octanol/water distribution coefficient of 4,200 compounds. The NCI database includes around 20,000 compounds and 60 prediction tasks from drug reaction experiments to clinical pharmacology studies. The free solvation database (FreeSolv) \cite{mobley2014freesolv} provides experimental and calculated hydration free energy of molecules in water. In our experiments, each dataset is split into training, validation and test subsets following the 8:1:1 ratio.

\subsubsection{Network Configuration}
In the experiments, the vertex attributes contain atom types, valence, formal charge, etc., and the edge attributes include bond types, same ring, etc. 
The GAAN architecture can be depicted as GAC(32)-PMF-GAC(64)-PMF-GAC(128)-PMF-GAC(256)-PMF-GMP-Tanh, where Tanh is the activation function. The GAAN model is trained with multitask cross entropy loss for no more than 200 epochs. When the molecular features are extracted, a multitask classifier is used to predict more than six hundred molecular property tasks. The learning rate starts from 0.001 and the batch size is also 64 empirically. Early stopping strategy is also adopted to obtain the best network model in training phase.

\subsubsection{Experimental Results}
To evaluate the classification capability of GCNNs, we compare them on four physiology datasets with more than six hundred classification tasks. We calculate the mean AUC-ROC scores over all of the tasks for each dataset as the metric for indicating the classification precision. According to the experimental results summarized in Table \ref{tab-classification}, the proposed GAAN model achieves the best performances on all the datasets. The performances of SCNN model have rooms to improve on the Tox21 dataset. The SCNN and NFP models show similar results on the ToxCast dataset, while the ChebNet model is superior to SCNN and NFP on all datasets. In particular, the AGCN model obtains an significant improvement compared to SCNN, NFP and ChebNet on datasets with small molecular graphs such as Tox21, ClinTox and ToxCast. On the SIDER dataset, AGCN achieves a slight boost because SIDER contains larger molecular graphs than other datasets. In short, the experiment results demonstrate the advantage of the GAAN model in classification tasks.

To evaluate the predictive power, we perform the regression tasks to compare the GAAN model with four state-of-the-art GCNNs. The Root Mean Square Error (RMSE) is adopted as the metric of regression tasks. The lower value indicates the better performance. Each experiment is repeated ten times to obtain a stable evaluation result and the standard deviation is calculated from statistical data for each dataset. Experimental results on the four physical chemistry datasets (ESOL, Lipophilicity,  NCI, and FreeSolv) presented in Table \ref{tab-regression} show that AGCN achieves different degrees of improvement compared to SCNN, NFP, and ChebNet. The NFP model receives a higher RMSE than other models on all datasets except Lipophilicity. The ChebNet model is not as good as others on Lipophilicity. By contrast, our GAAN model demonstrates the significant improvement compared to all other methods on all physical chemistry datasets.

Through the analysis of the experimental results, it is obvious that different design principle leads to different ability of extracting information from graphs. The spectral kernel in the SCNN model only connects one-hop neighbors. This over-localized kernel is unable to cover special structures in graphs such as functional groups in molecules. The ChebNet model extends the one-hop to the $K$-hop kernel and avoids over-localization compared to SCNN. But the kernel is still applied among graph structures. It is feasible when graphs share common sub-structures, such as $-COOH$ (carboxyl group) in molecules. However, ChebNet does not work well if there are large quantities of sub-structures classifying graphs into multiple categories, especially when the scale of graphs is small. This may be the reason why the ChebNet performs decently on large graphs such as the mean ROC-AUC 0.5914 on SIDER dataset, but dramatically worse on small graphs, e.g., RMSE 0.465 on ESOL and mean ROC-AUC 0.7573 on ClinTox. In addition, with the increase of graph scale, ACGN cannot learn more effective information from graphs as the results shown on SIDER. This indicates that the residual laplacian learning for hidden structures is more useful when data is short. As we illustrated above, GAC involves both global topology and local structure information, and more importantly, PMF is adept at aggregating the hierarchical information of graphs. In other words, the proposed PMF with GAC can promote GCCNs to learn features effectively from both small and large graphs.

\section{Conclusion}
\label{sec5}
In this paper, we introduce a graph convolutional network architecture, GAAN, in which PMF is devised to implement the aggregation strategy by iteratively abstracting the interior graph structures at different layers. The experiment results show that our strategy, emphasizing the importance of local sensitivity and capturing the interior graph structures naturally, bring us the improvements in the prediction of molecular property. This approach can be expended for other application for many fields, for example, taking the star structures into consideration for the requirements of social networks.  Moreover, many aspects of GCNN are worth exploring. The inference operations in GCCN can be divided and parts of them should be conducted in the early stages of graph convolution based on local graph structures. More mature techniques in the graph field into GCCNs, especially the frequent subgraph mining and subgraph isomorphism. New architectures of GCCN, which deeply combine the convolutional kernels and the aggregation methods according to the dynamic hypergraph structures.

\clearpage

\bibliographystyle{named}
\bibliography{ijcai19}

\end{document}